\documentclass[letterpaper, 10 pt, conference]{ieeeconf}  

\IEEEoverridecommandlockouts                              

\usepackage{graphics} 
\usepackage{epsfig} 
\usepackage{mathptmx} 
\usepackage{times} 
\usepackage{amsmath} 
\usepackage{amssymb}  
\usepackage{booktabs}
\usepackage{graphicx}

\usepackage{multirow}
\usepackage{hyperref}
\usepackage{color}

\title{\LARGE \bf
RT-Grasp: Reasoning Tuning Robotic Grasping via\\
Multi-modal Large Language Model
}

\author{
Jinxuan Xu$^{1*}$, Shiyu Jin$^{2}$, Yutian Lei$^{2}$, Yuqian Zhang$^{1}$ and Liangjun Zhang$^{2}$\\
\thanks{$^{1}$Jinxuan Xu and Yuqian Zhang are with Rutgers University, Department of Electrical and Computer Engineering. 
}%
\thanks{$^{2}$Shiyu Jin, Yutian Lei, and Liangjun Zhang are with Robotics and Autonomous Driving Lab (RAL), Baidu Research, USA.
}%
\thanks{$^*$Work done while the author was an intern at Baidu Research, USA.}
}

\begin{document}

\maketitle
\thispagestyle{empty}
\pagestyle{empty}

\begin{abstract}

Recent advances in Large Language Models (LLMs) have showcased their remarkable reasoning capabilities, making them influential across various fields. 
However, in robotics, their use has primarily been limited to manipulation planning tasks due to their inherent textual output.
This paper addresses this limitation by investigating the potential of adopting the reasoning ability of LLMs for generating numerical predictions in robotics tasks, specifically for robotic grasping.
We propose Reasoning Tuning, a novel method that integrates a reasoning phase before prediction during training, leveraging the extensive prior knowledge and advanced reasoning abilities of LLMs.
This approach enables LLMs, notably with multi-modal capabilities, to generate accurate numerical outputs like grasp poses that are context-aware and adaptable through conversations.
Additionally, we present the Reasoning Tuning VLM Grasp dataset, carefully curated to facilitate the adaptation of LLMs to robotic grasping.
Extensive validation on both grasping datasets and real-world experiments underscores the adaptability of multi-modal LLMs for numerical prediction tasks in robotics.
This not only expands their applicability but also bridges the gap between text-based planning and direct robot control, thereby maximizing the potential of LLMs in robotics.
More details and videos of this work are available on our project page: \textcolor{magenta}{\href{https://sites.google.com/view/rt-grasp}{https://sites.google.com/view/rt-grasp}}.

\end{abstract}

\begin{figure*}[ht]
    \centering
    \includegraphics[width=\textwidth]{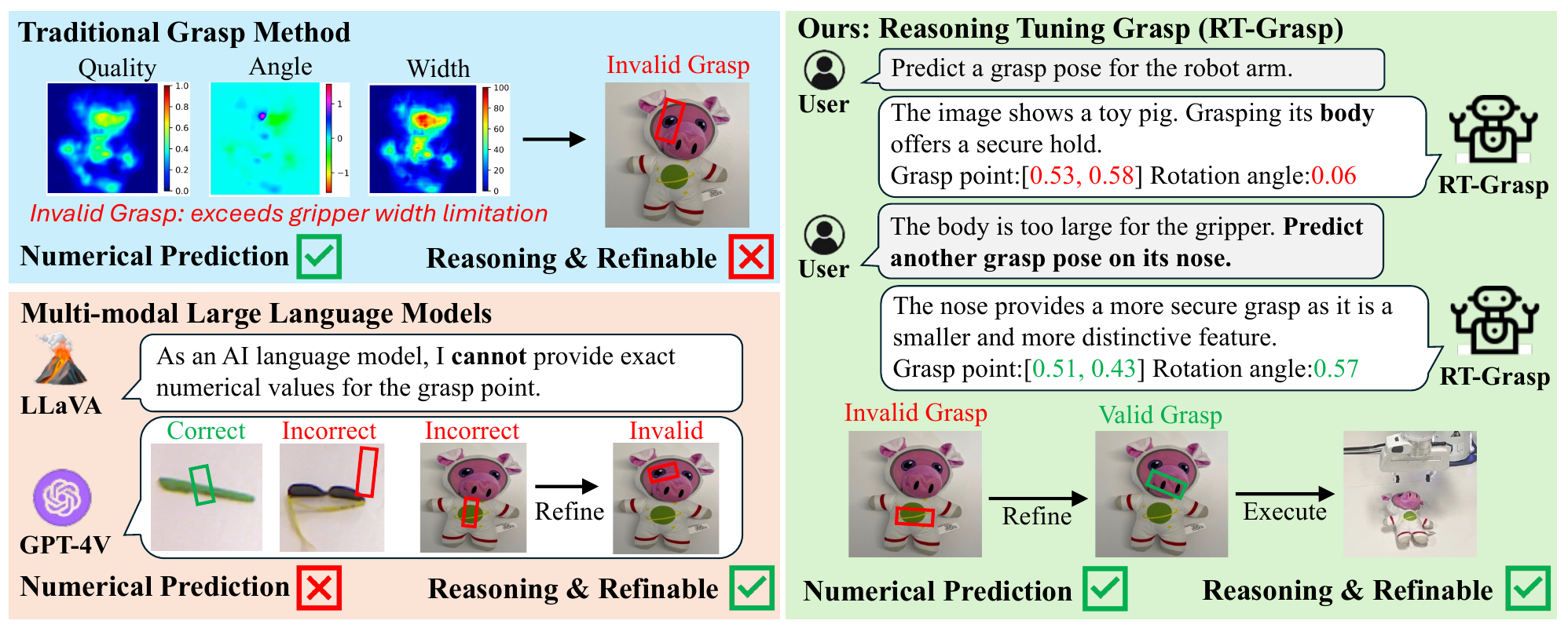}
    \caption{Comparing three robotic grasping approaches: 1) Traditional CNN-based algorithms produce fixed poses, which lack adaptability in practical situations. 2) Multi-model LLMs output adaptable grasping strategies but lack precise numerical predictions. 3) Ours combines the best of both, predicting adaptable numerical grasping informed by reasoned strategies.}
    \label{fig:1}
\end{figure*}

\section{INTRODUCTION}
The growth of artificial intelligence in recent years has been significantly driven by the emergence of large language models (LLMs). These models, packed with vast knowledge and advanced reasoning ability, have revolutionized our approach to various tasks, especially those involving language processing. In robotics, LLMs play a crucial role in facilitating direct interactions between robots and humans. For instance, in tasks such as robot manipulation planning, many studies \cite{lin2023text2motion, ahn2022can, huang2022inner} have utilized LLMs to interpret natural language commands from users and translate them into feasible multi-step plans for robots.
However, despite their potential in robotics, LLMs' application has predominantly been limited to such planning tasks. A notable bottleneck lies in the textual nature of LLM outputs, which often pose challenges for tasks requiring precise numerical outputs.

Recently, multi-modal LLMs have expanded LLM capabilities by understanding both text and images. In robotics, they bridge the gap between perception and planning, addressing a variety of embodied reasoning tasks \cite{driess2023palm, brohan2023rt}. However, their image understanding lacks precision, for example, they often struggle to accurately determine object locations, despite providing general descriptions.
Although models like GPT-4 with vision~\cite{achiam2023gpt} show promise in tasks like object detection, they encounter difficulties when tasked with making unique numerical predictions, such as grasp poses in robotic grasping (refer to Fig. \ref{fig:1}).
Another significant challenge in the application of multi-modal LLMs in robotics lies in the instability and verbosity of their textual outputs, which renders them unreliable for tasks requiring precise manipulation. While certain robotic tasks can benefit from the integration of multi-modal LLMs, their capacity for direct numerical prediction remains largely unexplored.

This paper investigates the potential use of multi-modal LLMs in numerical prediction tasks, specifically focusing on the domain of robotic grasping. Robotic grasping, considered one fundamental yet most challenging area in robotics, revolves around the generation of precise grasp poses essential for subsequent robot manipulation.

Traditional robotic grasping methods typically rely on deterministic predictions, which often fail in real-world scenarios due to their lack of reasoning capabilities. Most existing methods \cite{morrison2020learning, kumra2020antipodal}, using CNN-based architectures, excel in experimental accuracy on benchmark datasets, but struggle in practical applications. For example, these traditional models may produce theoretically correct predictions that prove impractical in execution, as shown in Fig. \ref{fig:1} and labeled as invalid. Such predictions are hard to apply across robot arms due to varying gripper constraints. Additionally, some theoretically correct grasps may result in unsafe actions, such as targeting the sharp ends of screwdrivers during grasping.

Hence, adopting a non-deterministic approach equipped with reasoning ability is crucial. This capability not only allows the model to generate practical grasp poses applicable across various settings but also allows the refinement of predictions based on user commands.
Here a question is posed: \textit{can the reasoning capabilities inherent in LLMs be utilized for numerical prediction tasks in robotics?} This paper offers a positive answer, showcasing an adaptation of multi-modal LLMs to robotic grasping tasks.

To efficiently utilize the reasoning capability of multi-modal LLMs for numerical predictions, we propose a novel methodology called Reasoning Tuning. This approach introduces a crucial reasoning phase preceding the numerical prediction step during training. The primary objective of this reasoning phase is to encourage the model to ground its predictions in logical reasoning principles. For instance, the model first logically infers attributes such as the object's type, shape, position, and a fundamental grasping principle. Subsequently, the numerical prediction is derived from this reasoning phase. This reasoning phase aims to unlock the valuable information encapsulated within multi-modal LLMs, leveraging their vast knowledge of general object attributes. 
In this paper, we empirically showcase that fine-tuning multi-modal LLMs with the integration of this reasoning phase enhances their efficacy in generating numerical predictions in robotic grasping.

We investigate two economical training strategies for the proposed Reasoning Tuning: pre-training and Low-Rank Adaptation (LoRA) fine-tuning \cite{hu2021lora}. Our intent behind this investigation is to present a more resource-efficient method for transferring the capabilities of multi-modal LLMs to downstream robotic tasks.

In summary, our work focuses on adapting multi-modal LLMs for numerical prediction tasks, specifically in the domain of robotic grasping. In contrast to deterministic traditional methods, our approach not only incorporates advanced reasoning capabilities but also introduces a novel paradigm for refining predictions, as illustrated in Fig. \ref{fig:1}. The main contributions can be summarized as follows:

\begin{itemize}
    \item We propose Reasoning Tuning, a novel methodology that utilizes the inherent prior knowledge of pre-trained multi-modal LLMs, facilitating their adaptation to tasks requiring numerical predictions.
    \item We present our dataset, Reasoning Tuning VLM Grasp dataset, designed for fine-tuning multi-modal LLMs for robotic grasping.
    \item We empirically validate the proposed method for robotic grasping using two computationally efficient training strategies and conduct real-world hardware experiments. Our results demonstrate its effectiveness and its ability to refine grasping predictions based on user commands.
    
\end{itemize}

\begin{figure*}[ht]
    \centering
    \includegraphics[width=\textwidth]{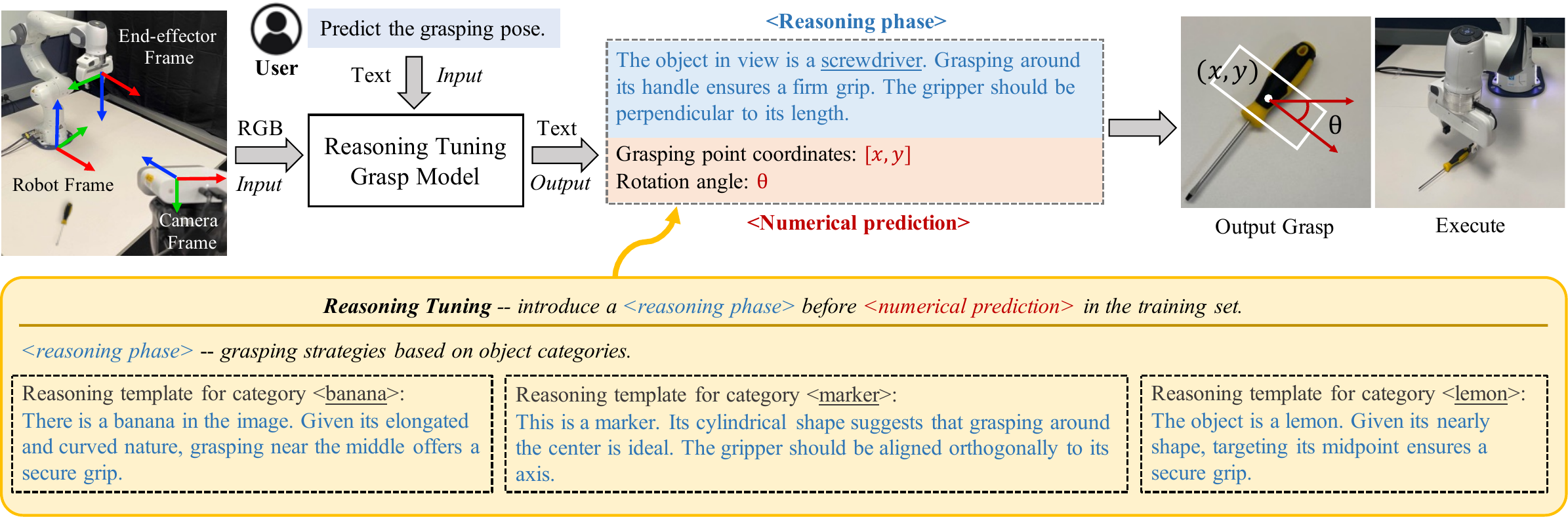}
    \caption{Overview. The proposed method processes RGB images and user instructions to yield text outputs, which comprise both a reasoning phase and a numerical grasp pose prediction $p=\{x,y,\theta\}$. The reasoning phase analyzes the object's shape and structure based on its category and generates corresponding grasping strategies. 
    }
    \label{fig:2}
\end{figure*}

\section{RELATED WORK}

\subsection{Robotic Grasping}
Traditionally, robotic grasping has heavily relied on analytical approaches~\cite{maitin2010cloth, domae2014fast, roa2015grasp}. These methods primarily focus on understanding object geometry or analyzing contact forces to determine a grasp that optimizes stability. However, these techniques often struggle to generalize well to unseen objects and can falter when confronted with irregularly shaped items.

In recent years, data-driven methods, particularly those leveraging convolutional neural networks (CNNs), have shown promising results~\cite{jiang2011efficient, lenz2015deep, pinto2016supersizing, mahler2017dex, kumra2020antipodal, morrison2020learning, zhu2022learn}. 
These approaches leverage extensive datasets of labeled grasping examples to train models capable of predicting grasp poses. Despite their success, these models often suffer from overfitting. They also lack the ability to reason about the usage, category, material, and other properties of objects beyond their shape. This limitation restricts their effectiveness in real-world scenarios, particularly when grasping objects with unusual shapes or those requiring special handling due to their material properties or intended use.

\subsection{Language Grounding for Robotics}
\subsubsection{Language-conditioned Robotic Manipulation}
In recent years, the integration of natural language into robotic manipulation has garnered significant interest.
Studies~\cite{xu2023joint, chen2021joint,ito2022integrated,hatori2018interactively} have explored grasp detection grounded following language instructions within cluttered scenes. \cite{rao2018learning} performs grasping prediction based on language descriptions of object properties. Building upon advancements in language models~\cite{devlin2018bert, radford2021learning}, recent studies \cite{stepputtis2020language, zheng2022vlmbench, shridhar2022cliport, goyal2023rvt, ma2023liv, brohan2022rt} have successfully grounded more flexible language instructions into long-horizon manipulation tasks. However, these approaches often require extensive demonstrations to master image-based policies.

\subsubsection{LLMs for Robotic Manipulation}
With the rise of LLMs, there has been a surge in research exploring their capabilities for robotic manipulation. Many studies~\cite{ahn2022can, huang2022inner, jin2023alphablock} have integrated LLMs into closed-loop planning frameworks, decomposing language-conditioned long-horizon tasks into multiple manageable steps. However, bridging the gap between language instructions and actions in robotics remains a challenge.
Additionally, some studies~\cite{singh2023progprompt, vemprala2023chatgpt, huang2023voxposer} have employed program-like specifications to prompt LLMs, melding planning and action using a predefined library of action functions. While intriguing, these methods often face limitations stemming from basic action functions and typically rely on additional perception models, leading to reduced system efficiency and flexibility.
Recent studies~\cite{brohan2023rt} have made progress in narrowing the planning-action gap by leveraging multi-modal LLMs. However, the method has high data and computational requirements, limiting their feasibility in real-world applications. In contrast, our approach capitalizes on the inherent knowledge embedded within LLMs to achieve precise numerical predictions in the realm of robotics, offering a promising alternative to existing methodologies.

\section{ROBOTIC GRASPING}

In this work, the robotic grasping problem is defined as finding an antipodal grasp, perpendicular to a planar surface, given an n-channel image and accompanying textual instructions. Similar to~\cite{morrison2018closing, kumra2020antipodal}, the grasp pose can be parameterized as $g = \{x, y, \theta, w\}$, where $(x, y)$ indicates the 2D coordinates signifying the center point of the grasp pose; $\theta$ denotes the rotation angle of the gripper compared to the horizontal axis; $w$ represents the width of the rectangular grasping box, corresponding to the width of the gripper. However, in many studies, the inclusion of $w$ within the predicted grasp pose $g$ is usually considered non-essential~\cite{kumra2017robotic}, due to variations in gripper width limitations.

To this end, our study, with its primary focus on probing the efficacy of LLMs in numerical prediction tasks, assumes $w$ equals the maximum width of the gripper. This paper defines the grasp pose as:
\begin{align}
    p = \{x, y, \theta\},
\label{eqa:grasp_p}
\end{align}
where $(x, y)$ coordinates are normalized by image width and image height respectively, and rotation angle $\theta$ is represented in radians scaling to $(-\frac{\pi}{2}, \frac{\pi}{2})$, as illustrated in Fig. \ref{fig:2}.

\section{RT-GRASP}
In this section, we introduce Reasoning Tuning for robotic grasping (RT-Grasp), a novel method designed to bridge the gap between the inherent text-centric nature of LLMs and the precise numerical requirements of robotic tasks. Its primary objective is to facilitate multi-modal LLMs for numerical prediction by leveraging their extensive encapsulated prior knowledge.

A pre-trained multi-model LLM, such as LLaVA~\cite{liu2023visual}, can be directly fine-tuned in a fully supervised manner when given the image and the text instruction. The model is trained by predicting each token in the text output sequentially. The proposed Reasoning Tuning introduces a structured text output, which includes a reasoning phase and a subsequent numerical prediction. We created our image-text dataset for robotic grasping, named Reasoning Tuning VLM (Visual Language Model) Grasp dataset, used for fine-tuning multi-modal LLMs. Additionally, we introduce a method that automatically generates such image-text datasets using GPT-3.5~\cite{ouyang2022training}, which can be applied to datasets for tasks beyond robotic grasping. Further details are presented in Section \ref{sec:reasoning_tuning}. Furthermore, we discuss two cost-efficient training strategies employed in our experiments in Section \ref{sec:framework}.

\subsection{Reasoning Tuning}
\label{sec:reasoning_tuning}

In this section, we introduce Reasoning Tuning, a method that fine-tunes multi-modal LLMs using image-text pairs as inputs and generating structured text outputs. This structured output comprises an initial reasoning phase followed by a subsequent numerical prediction, as illustrated in Fig. \ref{fig:2}. Notably, the entire output is in textual form, and the model is trained to predict corresponding tokens sequentially. By incorporating a reasoning phase at the outset of the output, we encourage the model to generate precise predictions based on logical reasoning specific to the task.

\begin{figure}[b]
\vspace{-.2in}
    \centering
    \includegraphics[width=0.49\textwidth]{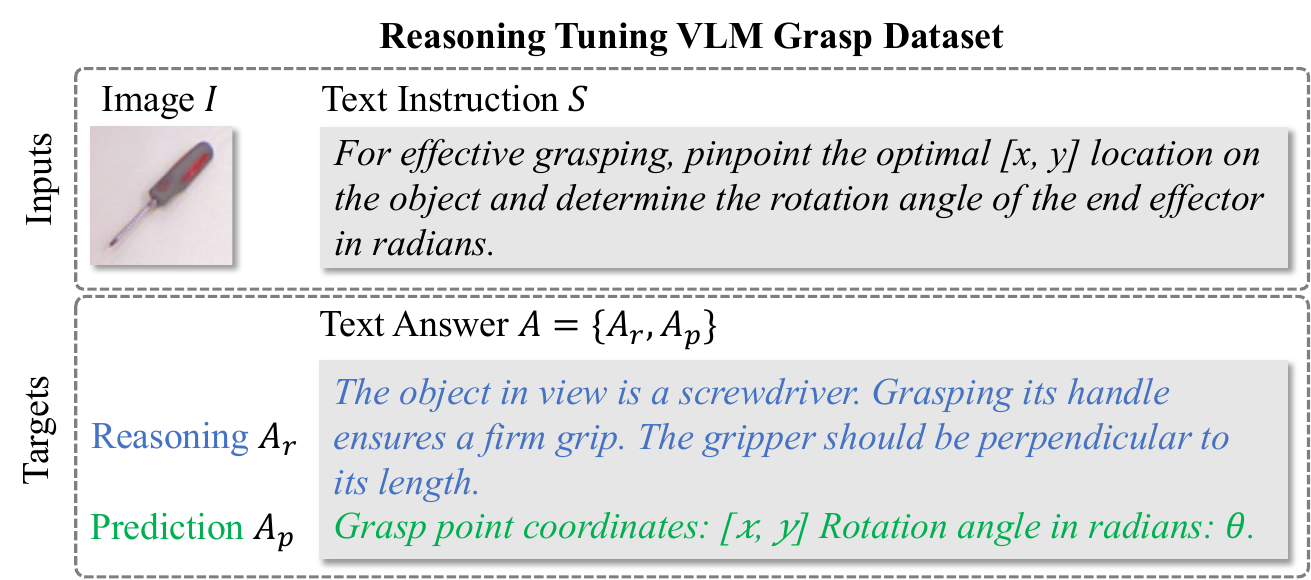}
    \vspace{-.2in}
    \caption{Illustration of a data sample from the Reasoning Tuning VLM Grasp dataset. The structured text answer contains both a reasoning phase and the ground truth grasp pose.
    }
    \label{fig:3}
\end{figure}

For robotic grasping, we created a new dataset for fine-tuning multi-modal LLMs, called the Reasoning Tuning VLM Grasp dataset. Each data sample includes an RGB image and a text instruction prompting the model to predict the grasp pose (refer to Fig. \ref{fig:3}). Additionally, the structured target text in this dataset contains a reasoning phase for the object within the input image, followed by a ground truth grasp pose. The reasoning phase provides a general description of the object, covering aspects such as shape and position, and suggests a corresponding grasping strategy. For instance, consider cups, which may vary in color, design, or material, but a general grasping strategy for them is universal by targeting the handle or the upper edge. Integrating such a reasoning phase guides the model to establish a broad understanding of the object and relevant grasping strategies, thereby facilitating a more informed numerical prediction in subsequent steps.

Existing datasets for robotic grasping typically comprise solely images and numerical ground truth grasp poses. In contrast, our Reasoning Tuning VLM dataset provides image-text pairs tailored specifically for integrating multi-modal LLMs into robotic grasping. In this dataset, images are sourced from the benchmark Cornell Grasp dataset~\cite{jiang2011efficient}, while the accompanying structured texts consist of a reasoning phase followed by ground truth grasp poses presented in textual format. Next, we detail the methodology employed to automatically generate corresponding texts in our dataset.

For the reasoning phase in the structured text, we generated templates based on object categories, as grasping strategies for objects of the same type are usually similar.
For each category, we create a series of different reasoning templates. In the structured text of each data sample, one reasoning template is randomly selected based on the object category, followed by the appending of the ground truth grasp pose in textual form (refer to Fig. \ref{fig:3}).

\begin{figure}[!t]
    \centering
    \includegraphics[width=0.49\textwidth]{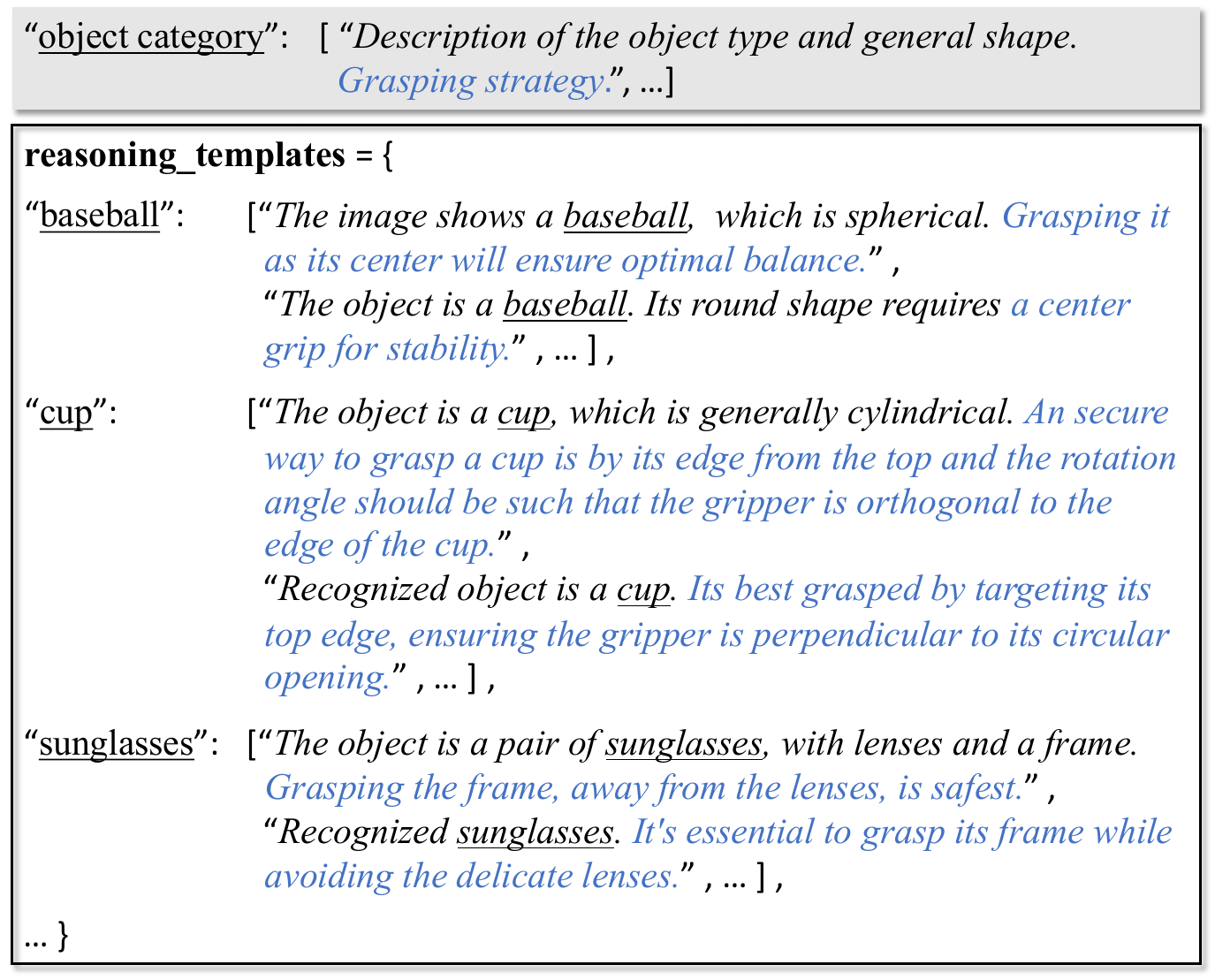}
    \caption{Examples of reasoning templates within the Reasoning Tuning VLM dataset.
    }
    \label{fig:templates}
\end{figure}

To ensure the quality of these reasoning templates, we adopt a multi-step approach. Initially, we prompt GPT-3.5~\cite{ouyang2022training} to generate a collection of templates tailored to each category. Subsequently, we instruct it to refine these drafts, removing redundant or irrelevant sentences. Finally, as a quality check, we manually verify the correctness and relevance of the generated templates. These reasoning templates typically describe the shape of the object and offer a general grasping strategy.
We present some examples of reasoning templates in Fig. \ref{fig:templates}, and the full collection and GPT-3.5 prompts can be found on our project page.

For the input text instruction in our dataset, we also employ GPT-3.5 to generate a series of consistent instruction templates pertaining to the robotic grasping task, and an example template is presented in Fig. \ref{fig:3}. Notably, the methodology behind creating this image-text dataset is adaptable to other numerical prediction tasks beyond robotic grasping. Adjusting the strategies in the reasoning phase can draw upon the appropriate prior knowledge embedded within LLMs tailored for different tasks.

\begin{figure*}
    \centering
    \includegraphics[width=\textwidth]{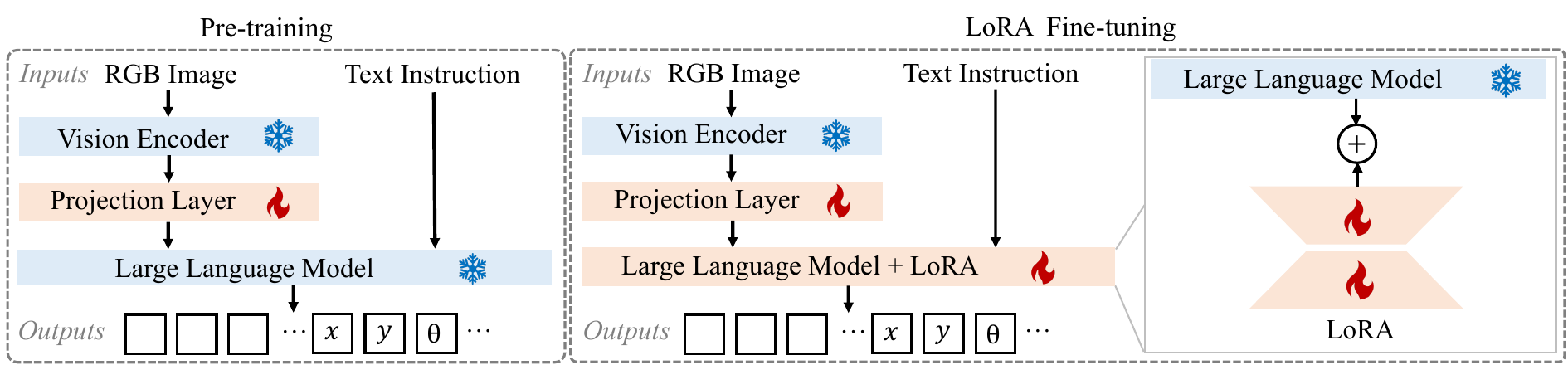}
    \caption{Two training strategies. 1) Pre-training: only parameters of the projection layer are trainable; 2) LoRA fine-tuning: only parameters of the projection layer and LoRA model are trainable.}
    \label{fig:4}
\end{figure*}

\subsection{Training Strategy}
\label{sec:framework}
In our dataset, for each image $I$, we have a single round conversation data form $(S, A)$, where $S$ represents the input instruction and $A$ is the associated target answer.
This paper performs two training strategies: pre-training and LoRA fine-tuning, as illustrated in Fig. \ref{fig:4}. Both strategies utilize an auto-regressive training objective following LLaVA~\cite{liu2023visual}. To elaborate, for a sequence of length $l$, the probability of producing the target answer $A$ is formulated as
\begin{align}
    p(A|I,S) = \prod^{l}_{i=1}p_{\theta_m}(a_i|I,S,A_{<i}),
\label{eqa:objective}
\end{align}
where $\theta_m$ is trainable parameters in the model; $a_i$ represents the current prediction token; $A_{<i}$ indicate answer tokens before the current token $a_i$. 

In our Reasoning Tuning VLM dataset, we define the target answer $A=\{A_r, A_p\}$, which consists of two phases: $A_r$ representing the texts for the reasoning phase, and $A_p$ indicating the predictions of grasp poses including coordinates $[x, y]$ and rotation angle $\theta$. It is worth noting that since these numerical predictions $A_p$ essentially are also in text format, they are generated as tokens first by LLMs and then converted to textual numbers. Then \autoref{eqa:objective} can be rewritten as
\begin{equation}
\begin{aligned}
    &p(A|I,S) = p(A_r|I,S) \cdot p(A_p|I,S,A_r)\\
    =\prod^{|A_r|}_{i=1} &p_{\theta_m} (a_i|I,S,A_{r_{<i}}) \cdot  \prod^{|A_p|}_{j=1}p_{\theta_m}(a_j|I,S,A_r,A_{p_{<j}})
\end{aligned}
\end{equation}
where $p(A_r|I,S)$ denotes the probability of producing the reasoning texts, and $p(A_p|I,S,A_r)$ is the probability of producing grasp pose predictions conditioned on the input image $I$, instruction $S$, and reasoning phase texts $A_r$. The total length of the entire textual answer $A$ is $l=|A_r|+|A_p|.$

\subsubsection{Pre-training}
Within this training strategy, both the visual encoder and weights of the LLM are maintained in a frozen state. Only weights of the projection layer, which aligns image features with the word embedding space of the LLM, are updated.

\subsubsection{LoRA Fine-tuning}
To further enhance the performance, we adopt LoRA \cite{hu2021lora} fine-tuning, a computationally efficient technique that adds an external model to the existing LLM. Specifically, we inject LoRA into all linear layers within the LLMs. Notably, both the vision encoder and the original LLM remain frozen. Only weights of added LoRA and the projection layer are set as trainable parameters.

\section{Experiments}

In this section, we assess the performance of the proposed approach using both grasping datasets (Section \ref{sec:exp_cornell}) and household test objects on real robots (Section \ref{sec:exp_real}).
Moreover, we have developed two additional variants of the Reasoning Tuning VLM Grasp dataset for an ablation study. This study underscores the enhanced performance achieved by introducing the reasoning phase.

\subsection{Evaluation on Reasoning Tuning VLM Grasp datasets}
\label{sec:exp_cornell}

\subsubsection{Setup}
For all experiments, we utilize LLaVA-7B-v0 \cite{liu2023visual} as the base model, which is derived from the large language model LLaMA-7B \cite{touvron2023llama}. For the vision encoder, we employ the CLIP ViT-L/14 \cite{radford2021learning} to extract image features. During the pre-training, we set the batch size to $32$ with a learning rate of $2\times 10^{-3}$. During the LoRA fine-tuning, the batch size remains $32$ and the learning rate is $5\times 10^{-4}$. And we choose a rank $r=64$ and $\alpha=32$ for LoRA configurations.

\subsubsection{Datasets}
We evaluate the proposed method using our Reasoning Tuning VLM Grasp dataset. This dataset sources its RGB images from the benchmark Cornell Grasp dataset \cite{jiang2011efficient}, which consists of $885$ images representing $240$ distinct objects.
We have manually divided these objects into $74$ different categories, formulating specific grasping strategies for each category as introduced in Section \ref{sec:reasoning_tuning}. Given the relatively limited number of images, we have implemented data augmentation techniques such as image rotation, zooming, and random cropping, by following related studies \cite{kumra2020antipodal, kumra2022gr, zhang2023htc}. Consequently, we have $76$k image-text paired grasp samples, and only positively labeled grasps were included.

\subsubsection{Evaluation metrics}
We follow a cross-validation setup as in previous works and partition the datasets into $5$ folds. Both image-wise and object-wise splits are utilized for evaluation. 
Performance is reported using the rectangle metric \cite{jiang2011efficient}. A grasp pose is deemed valid if fulfills the following two conditions:
\begin{itemize}
    \item The Intersection over Union (IoU) score between the predicted and target rectangles exceeds 25\%.
    \item The angular deviation between the orientations of the predicted and target rectangles is less than $30$ degrees.
\end{itemize}
This metric requires a grasp rectangle representation, while our method predicts the grasp pose without the width $w$. Thus, to evaluate the accuracy, we convert the grasp pose $p$ combined with the ground truth $w$ into the rectangle representation.

\subsubsection{Ablation studies}
\begin{figure}[ht]
\vspace{-.1in}
    \centering
    \includegraphics[width=0.49\textwidth]{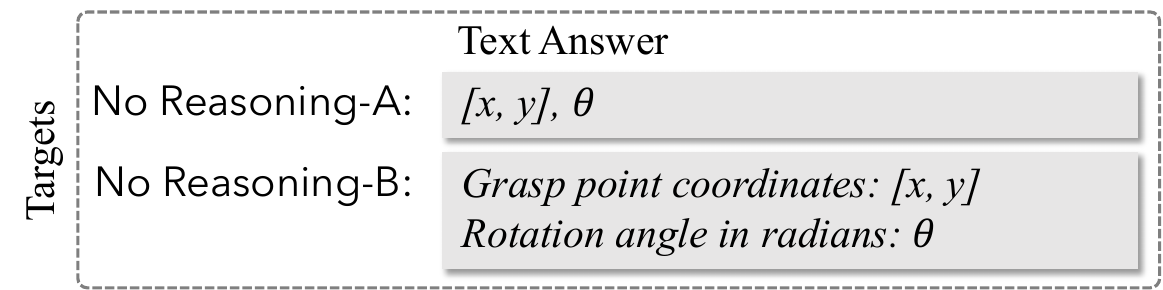}
    \caption{Illustration of two variants in ablation studies.
    }
    \label{fig:ablation}
\end{figure}

To demonstrate the effectiveness of integrating the reasoning phase, we have developed two variants of the structured text answer in our Reasoning Tuning VLM Grasp dataset. Neither variant includes the reasoning phase in their text answers, as shown in Fig. \ref{fig:ablation}. These variants maintain the image and input text instructions identical to those in the Reasoning Tuning VLM Grasp dataset, differing only in the text answers:
\begin{itemize}
    \item \texttt{No Reasoning-A}: This variant solely contains the textual grasp pose $p$.
    \item \texttt{No Reasoning-B}: This variant enhances the textual grasp pose with simple prompts. For instance, it specifies that $(x, y)$ denotes the center point coordinates, and $\theta$ indicates the rotation angle.
\end{itemize}

\subsubsection{Results}

\begin{table}[!b]
\caption{Results on grasping datasets. \textit{GR-ConvNet results are from \cite{kumra2020antipodal}.}}
    \renewcommand{\arraystretch}{1.3}
    \centering
    \setlength{\tabcolsep}{3.8mm}
    \begin{tabular}{l|cc}
    \hline
        Method & \multicolumn{2}{c}{Grasp Accuracy (\%)} \\ 
        & \multicolumn{1}{c}{Image-Wise (IW)} & \multicolumn{1}{c}{Object-Wise (OW)}   \\
        \hline
        \addlinespace[1ex]
        \multicolumn{3}{l}{\textit{Deterministic traditional method on Cornell Grasp dataset~\cite{jiang2011efficient}}} \\
        \hline
        GR-ConvNet \cite{kumra2020antipodal} & 96.60 & 95.50 \\
        \hline
        \addlinespace[1ex]
        \multicolumn{3}{l}{\textit{Pre-training on RT VLM Grasp dataset} (mean$\pm$std)} \\
        \hline
        No Reasoning-A & 65.70$\pm$0.87 & 61.55$\pm$1.32 \\
        No Reasoning-B  & 72.94$\pm$2.08 & 67.04$\pm$3.46 \\
        Ours (RT-Grasp) & 74.41$\pm$0.88 & 72.61$\pm$2.78 \\
        \hline
        \addlinespace[1ex]
        \multicolumn{3}{l}{\textit{LoRA Fine-tuning on RT VLM Grasp dataset} (mean$\pm$std)} \\
        \hline
        No Reasoning-A & 58.44$\pm$6.04 & 50.31$\pm$14.34 \\
        No Reasoning-B & 69.15$\pm$11.00 & 67.44$\pm$9.99 \\
        Ours (RT-Grasp) & \textbf{84.05}$\pm$0.78 & \textbf{77.02}$\pm$0.93 \\
        \hline
    \end{tabular}
    \label{tab:accu}
\end{table}

\begin{figure}[!b]
    \centering
    \includegraphics[width=0.49\textwidth]{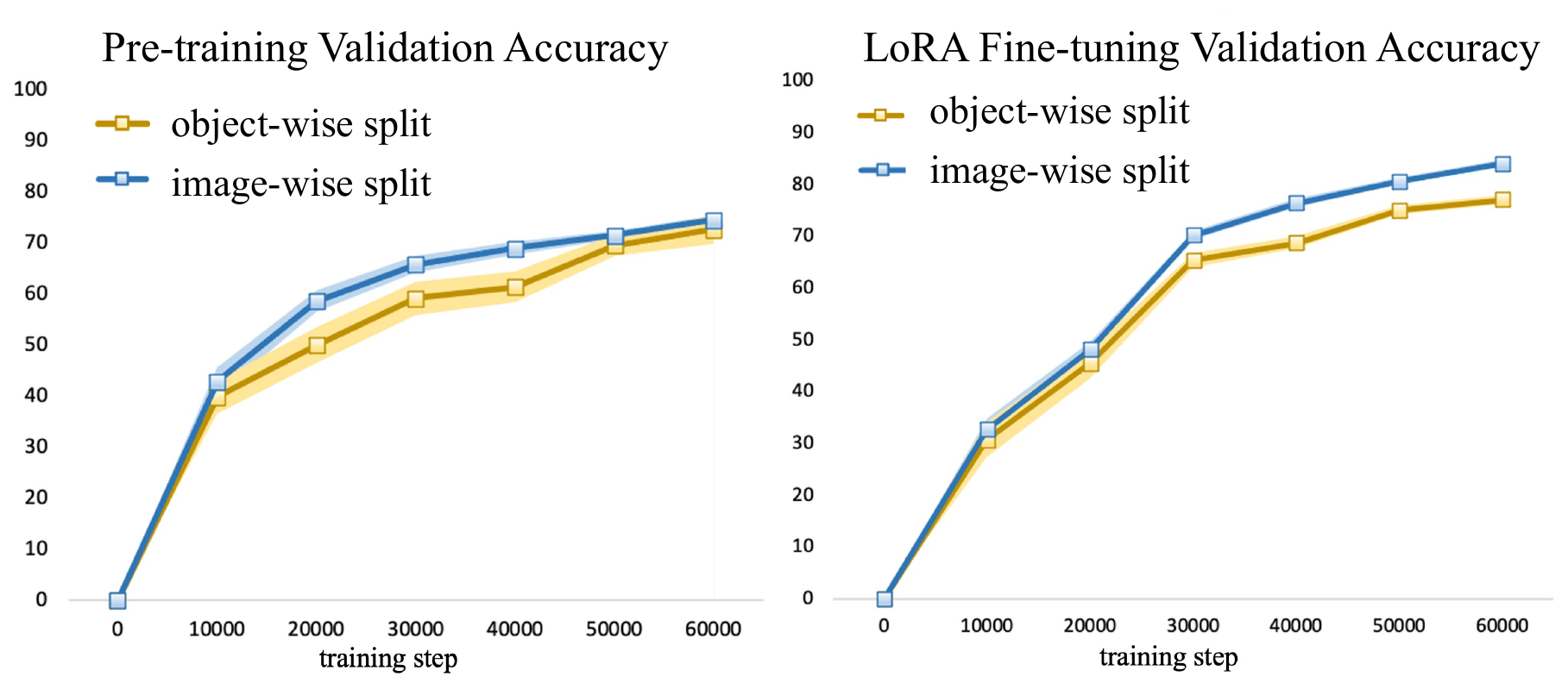}
    \caption{Validation accuracy of our method (RT-Grasp) for two training strategies.}
    \label{fig:5}
\end{figure}

In Table \ref{tab:accu}, we present the grasp prediction accuracy of two training strategies on our Reasoning Tuning VLM Grasp dataset. Given that our dataset originates from images of the Cornell Grasp dataset, we also include results from traditional grasping algorithms applied to this dataset. Notably, our reasoning-tuned model directly generates precise numerical values, whereas the traditional grasping method outputs heatmaps requiring post-processing to derive grasp poses.

Our primary objective is to develop a method leveraging the reasoning ability of multi-modal LLMs for numerical robotic tasks. As presented in Table \ref{tab:accu}, our method demonstrates promising grasping accuracy by incorporating the reasoning phase, underscoring the potential of multi-modal LLMs in numerical prediction.

Furthermore, the performance of \texttt{No Reasoning-A}, representing a straightforward adaptation of a language model to numerical prediction, is subpar. While the other variant \texttt{No Reasoning-B}, which incorporates simple prompts, shows an improvement over \texttt{No Reasoning-A}, its performance remains unsatisfactory. Our reasoning tuning method enhances accuracy by $8.7-26.7\%$ across all settings compared to \texttt{No Reasoning-A} and by $1.5-14.9\%$ compared to \texttt{No Reasoning-B}, highlighting the effectiveness of integrating the reasoning phase.
The validation accuracy curve of our method (RT-Grasp) throughout the training process is illustrated in Fig. \ref{fig:5}. Our approach effectively bridges the gap between planning and action in robotics, coupled with multi-modal LLMs' reasoning ability.

\subsection{Evaluation on real-world experiments}
\label{sec:exp_real}
\subsubsection{Setup}
In our real-world experiments, we conducted a comprehensive evaluation of both the traditional method GR-ConvNet~\cite{kumra2020antipodal} and our proposed approaches. We utilized a 7-DoF Franka Emika Panda equipped with a Franka Hand parallel gripper along with one Azure Kinect camera for grasping tasks. Visual input for all methods was a $400\times 400$ image crop of the workspace.

\subsubsection{Household test objects}

\begin{figure}[hb]
\vspace{-.1in}
    \centering
    \includegraphics[width=0.482\textwidth]{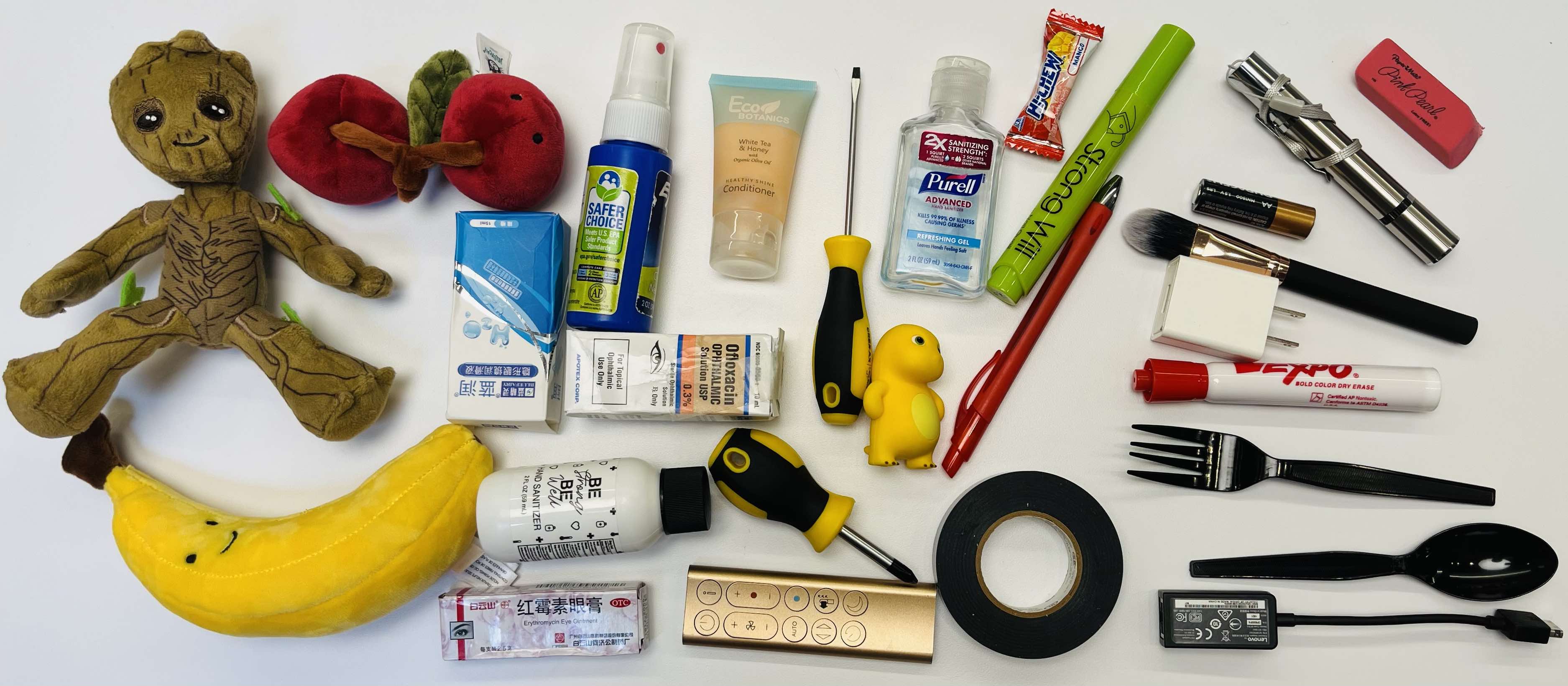}
    \caption{Household test objects for real-world grasping experiments.}
    \label{fig:real_dataset}
\end{figure}

Fig. \ref{fig:real_dataset} displays a diverse set of $27$ household objects selected for real-world experiments. Notably, all objects were unseen during training. Each object underwent testing in $5$ distinct positions and orientations in our experiments, totaling $135$ grasp tests for each method.

\subsubsection{Results}

\begin{figure}[!t]
    \centering
    \includegraphics[width=0.49\textwidth]{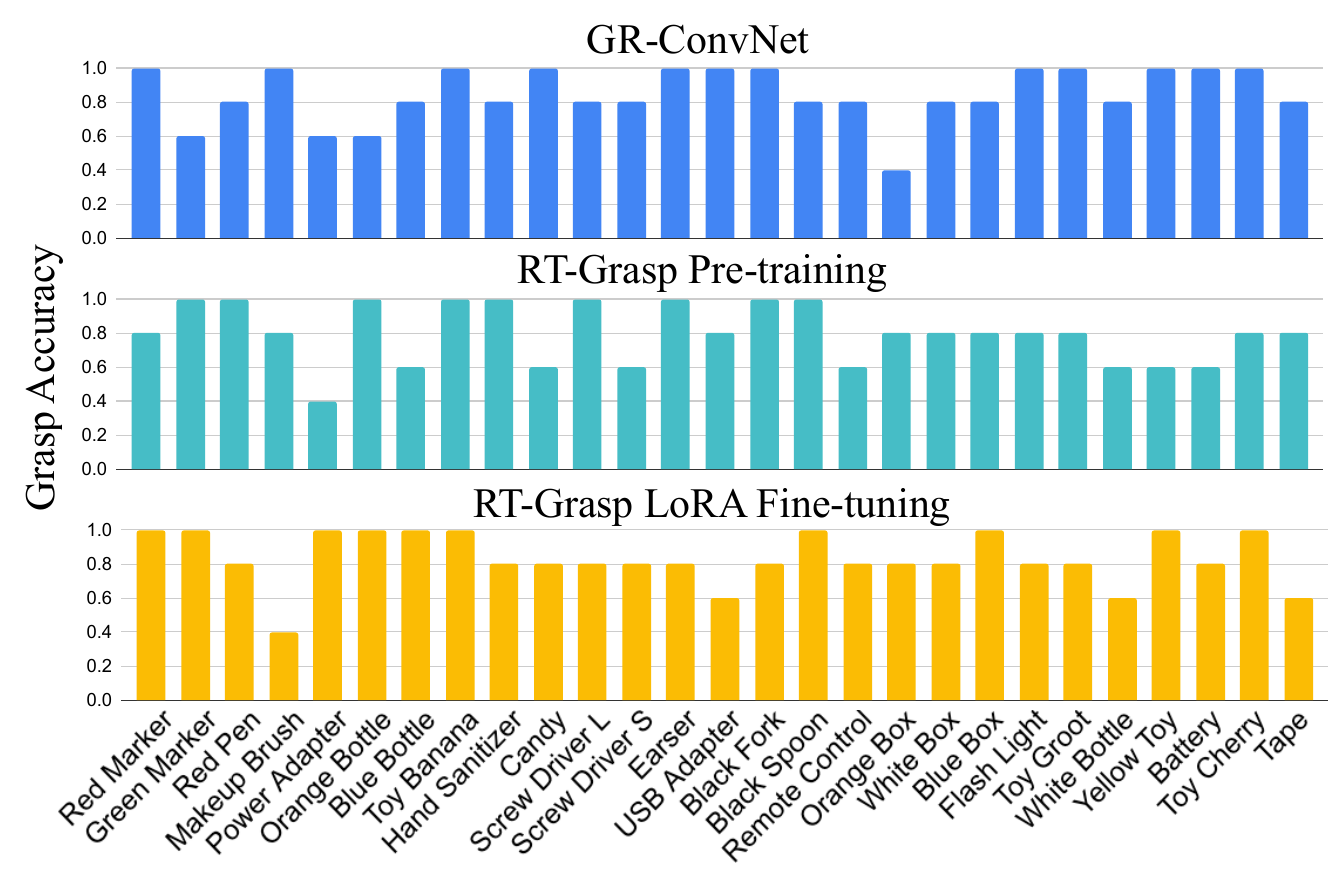}
    \caption{Grasp Accuracy on household test objects.}
    \label{fig:accu_hardware}
\end{figure}

We present grasp accuracy for both the traditional grasping method and our methods trained using two different strategies in Table \ref{tab:realworld}. The results showcase the competitive performance of our approach in real-world experiments compared to traditional methods. Notably, our method (RT-Grasp Pre-training) performs even better in real-world scenarios than on the grasping dataset, highlighting the effectiveness of our approach in adapting to unseen objects. Additionally, in Fig. \ref{fig:accu_hardware}, we present grasp success rates for 27 household test objects, illustrating that our methods successfully grasp all objects at least once during testing. This demonstrates the robustness of our approach across various object shapes.

Here we draw attention to the performance gap between grasping dataset accuracy (Table \ref{tab:accu}) and real-world experimental results (Table \ref{tab:realworld}). In real-world experiments, GR-ConvNet achieved an $85.19\%$ success rate, marking a noticeable degradation compared to its performance on the Cornell Grasp dataset. This underscores its limitations in effectively reasoning about unseen object attributes. Conversely, our proposed RT-Grasp achieved success rates of $80.00\%$ and $83.70\%$ for two training strategies respectively, closely aligning with its performance on VLM grasping datasets. These results demonstrate competitive accuracy in grasping household objects and emphasize the model's ability to generalize to unseen objects, including those from unseen categories such as unique toys. Additionally, in this testing, we utilize the initial predictions from RT-Grasp without any subsequent refinement, highlighting the effectiveness of adapting multi-modal LLMs to numerical robotic grasping tasks.

\begin{table}[!t]
\caption{Results on real-world evaluation.}
    \renewcommand{\arraystretch}{1.3}
    \centering
    \setlength{\tabcolsep}{7mm}
    \begin{tabular}{l|c}
        \hline
        Method & Grasp Accuracy ($\%$) \\ 
        \hline
        GR-ConvNet \cite{kumra2020antipodal} & 85.19 \quad (115/135) \\ 
        RT-Grasp Pre-training & 80.00 \quad  (108/135) \\ 
        RT-Grasp LoRA Fine-tuning & 83.70 \quad  (113/135) \\ 
        \hline

    \end{tabular}
    \label{tab:realworld}
\end{table}

\subsection{Interactive refinement and reasoning}
\label{sec:interact}

\begin{figure}[!t]
    \centering
    \includegraphics[width= 0.49\textwidth]{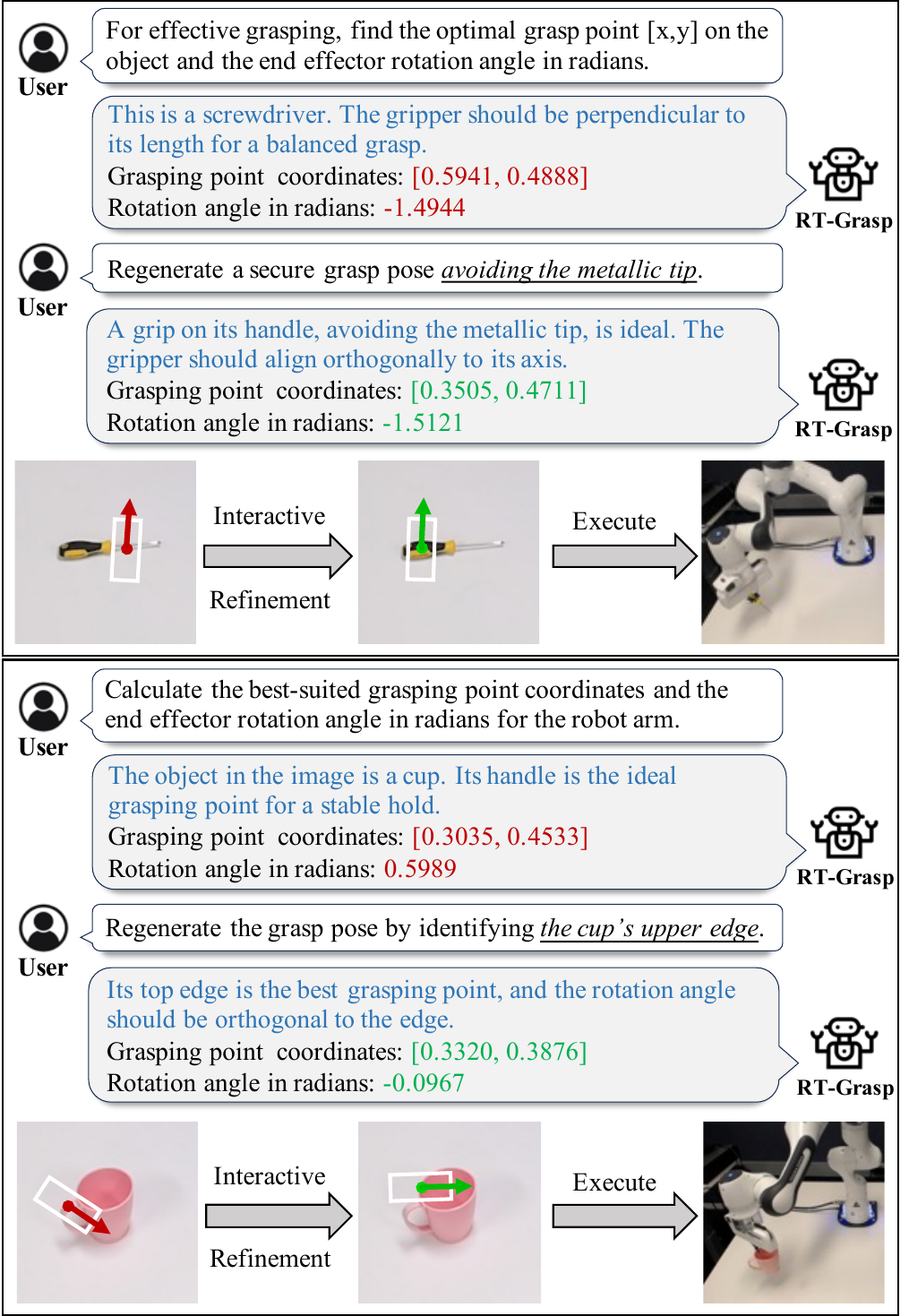}
    \caption{Reasoning and interactive refinement. Outputs from RT-Grasp include a reasoning phase (in blue) and a numerical grasp pose. The initial predicted grasp is indicated in red, while the grasp after refinement is denoted in green.}
    \label{fig:6}
\end{figure}

In contrast to deterministic traditional methods in robotic grasping, one notable advantage of adopting multi-modal LLMs is their capacity for the refinement of numerical outputs based on user instructions. Traditional CNN-based methods typically generate predictions solely based on input images, lacking the flexibility for refinement, thus limiting their applicability in real-world scenarios. However, our approach, enhanced with reasoning capabilities, offers the flexibility to generate different predictions in response to user commands.

In this section, we demonstrate the effectiveness of our model in reasoning by showcasing its ability to propose novel grasping strategies for unseen object categories. Additionally, our model exhibits the flexibility to refine grasp poses through real-time interaction with users.
The case depicted in Fig. \ref{fig:1} exemplifies our model's proficiency in generating innovative grasping strategies for objects from categories not encountered during training.
Furthermore, Fig. \ref{fig:6} illustrates two instances of dynamic interaction between users and the model, showcasing the model's ability to provide adaptable and refinable grasp predictions through multiple rounds of conversation.


\section{CONCLUSIONS}
This research underscores the potential of LLMs beyond their conventional text-centric applications. Our proposed method utilizes the extensive prior knowledge of LLMs for numerical predictions, specifically in robotic grasping. Through comprehensive experiments conducted on both benchmark datasets and real-world scenarios, we have demonstrated the efficacy of our approach.
For future work, we plan to extend the validation of our method by applying it to grasping datasets featuring a broader array of objects, such as the Jacquard dataset~\cite{depierre2018jacquard}. Moreover, the adaptation of multi-modal LLMs for numerical predictions in other robotic manipulation tasks is also a promising research direction.


\bibliographystyle{IEEEtran}
\bibliography{main}

\end{document}